%% file: main.tex
\documentclass[10pt,twocolumn,letterpaper]{article}

\usepackage{wacv}
\usepackage{times}
\usepackage{epsfig}
\usepackage{graphicx}
\usepackage{amsmath}
\usepackage{amssymb}
\usepackage{booktabs}

\def\wacvPaperID{71} 

\wacvalgorithmstrack   

\wacvfinalcopy 

\usepackage[caption=false]{subfig}
\usepackage{tabularx}
\usepackage{multirow}

\newcommand{\parsection}[1]{\vspace{1mm}\noindent\textbf{#1 }}

\ifwacvfinal
\usepackage[breaklinks=true,colorlinks,bookmarks=false]{hyperref}
\else
\usepackage[pagebackref=true,breaklinks=true,colorlinks,bookmarks=false]{hyperref}
\fi

\begin{document}

\title{Composite Learning for Robust and Effective Dense Predictions}

\author{Menelaos Kanakis$^{1}$\quad
    Thomas E. Huang$^{1}$\quad
    David Bruggemann$^{1}$\quad
    Fisher Yu$^{1}$\quad
	Luc Van Gool$^{1,2}$ \vspace{2mm} \\
$^1$ETH Z\"urich \quad $^2$KU Leuven
}

\maketitle
\thispagestyle{empty}

\input{text/abstract}
\input{text/intro}
\input{text/relwork}
\input{text/method}
\input{text/exp}

\input{text/conclusion}

\newpage

\renewcommand\thefigure{S.\arabic{figure}}    
\renewcommand\thetable{S.\arabic{table}}   

\begin{center}
	\textbf{\Large Supplementary Material}
\end{center}
\appendix

\input{text/additional_depth}
\input{text/additional_semseg}
\input{text/mtl}

\input{text/exp_details}

{\small
\bibliographystyle{ieee_fullname}
\bibliography{egbib}
}

\end{document}

%% file: text/abstract.tex
\begin{abstract}
Multi-task learning promises better model generalization on a target task by jointly optimizing it with an auxiliary task. 
However, the current practice requires additional labeling efforts for the auxiliary task, while not guaranteeing better model performance.
In this paper, we find that jointly training a dense prediction (target) task with a self-supervised (auxiliary) task can consistently improve the performance of the target task, while eliminating the need for labeling auxiliary tasks. 
We refer to this joint training as Composite Learning (CompL). 
Experiments of CompL on monocular depth estimation, semantic segmentation, and boundary detection show consistent performance improvements in fully and partially labeled datasets. 
Further analysis on depth estimation reveals that joint training with self-supervision outperforms most labeled auxiliary tasks. 
We also find that CompL can improve model robustness when the models are evaluated in new domains.
These results demonstrate the benefits of self-supervision as an auxiliary task, and establish the design of novel task-specific self-supervised methods as a new axis of investigation for future multi-task learning research.
\end{abstract}

%% file: text/intro.tex
\section{Introduction}

Learning robust and generalizable feature representations have enabled the utilization of Convolutional Neural Networks (CNNs) on a wide range of tasks.
This includes tasks that require efficient learning due to limited annotations.
A commonly used paradigm to improve generalization of target tasks is Multi-Task Learning (MTL), the joint optimization of multiple tasks. 
MTL exploits domain information contained in the training signals of related tasks as an inductive bias in the learning process of the target task~\cite{caruana1997multitask,caruana1998dozen}. 
The goal is to find joint representations that better explain the optimized tasks.
MTL has demonstrated success in tasks such as instance segmentation~\cite{dai2016instance} and depth estimation~\cite{chen2019towards}, amongst others.
In reality, however, such performance improvements are not common when naively selecting the jointly optimized tasks~\cite{kokkinos2017ubernet}.
To complicate things further, the relationship between tasks for MTL is also dependent on the learning setup, such as training set size and network capacity~\cite{standley2020tasks}.
As a consequence, MTL practitioners are forced to iterate through various candidate task combinations in search of a synergetic setting. 
This empirical process is arduous and expensive since annotations are required \emph{a priori} for each candidate task.

\begin{figure}
 \centering
  \subfloat[Multi-Task Learning]{\label{fig:MT_learning}
      \includegraphics[width=0.19\textwidth]{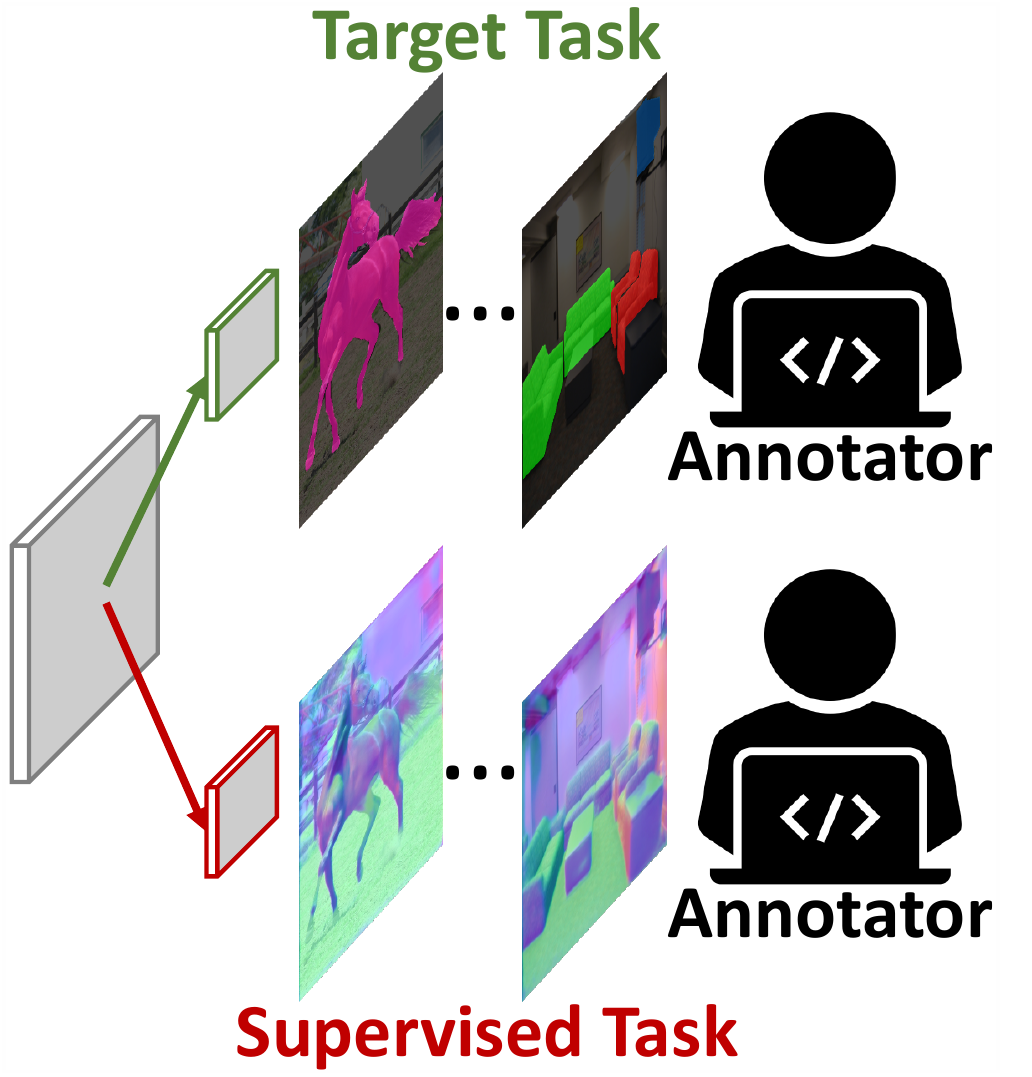}}
~
      \subfloat[Composite Learning (ours)]{\label{fig:CompL}
      \includegraphics[width=.19\textwidth]{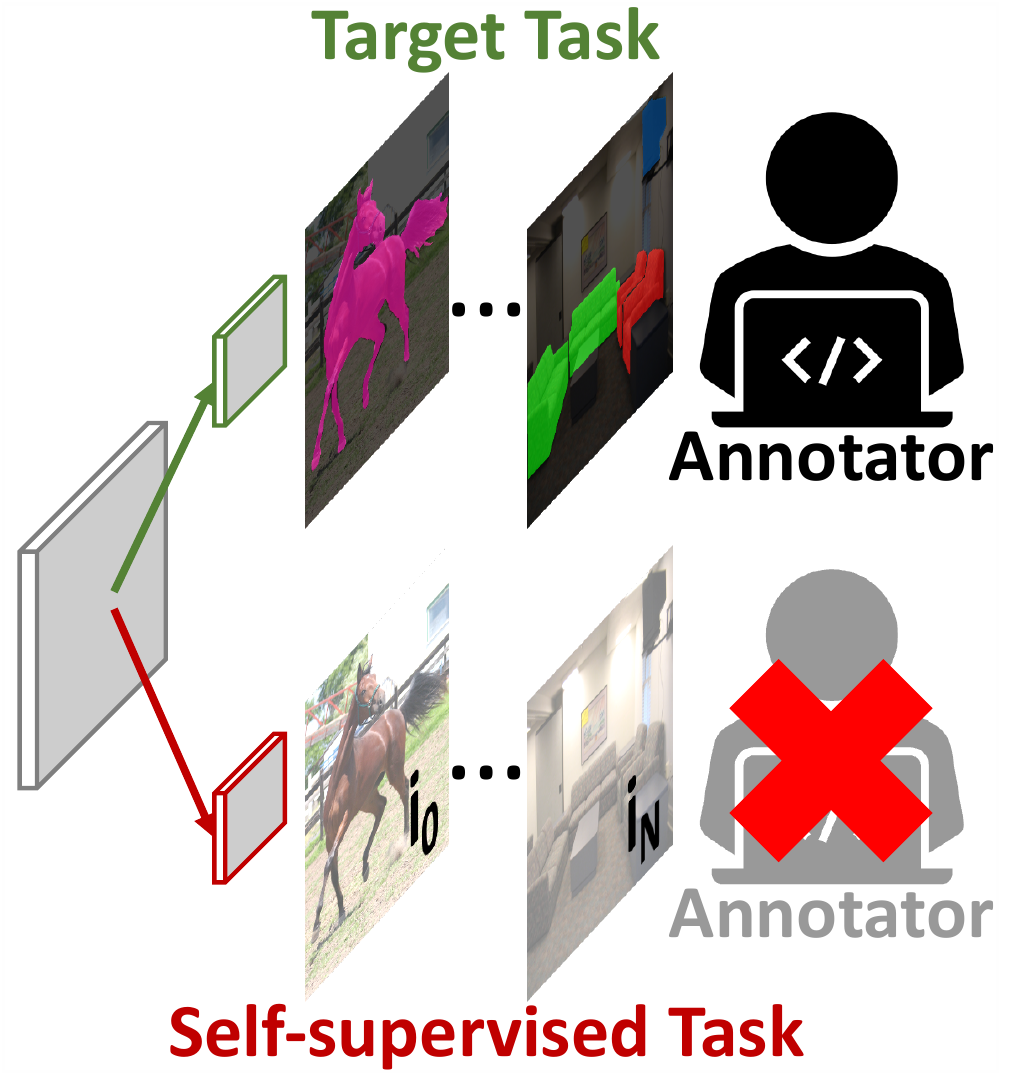}}
\caption{The generalization of target tasks can be improved by jointly optimizing with a related auxiliary task. 
(a) In traditional multi-task learning, one uses labeled auxiliary tasks that require manual annotation efforts. 
(b) In this paper, we show that jointly training a dense task with a self-supervised task can consistently improve the performance, while eliminating the need for additional labeling efforts.
}
\vspace{-0.2in}
\label{fig:overview}
\end{figure}

In this paper, we find that the joint optimization of a dense prediction (target) task with a self-supervised (auxiliary) task improves the performance on the target task, outperforming traditional MTL practices. 
We refer to this joint training as Composite Learning (CompL), inspired by material science where two materials are merged to form a new one with enhanced properties. 
The benefits and intuition of CompL resemble those of traditional MTL, however, CompL exploits the label-free supervision of self-supervised methods.
This facilitates faster iterations through different task combinations, and eliminates manual labeling effort for auxiliary tasks from the process.

We provide thorough evaluations of CompL on three dense prediction target tasks with different model structures, combined with three self-supervised auxiliary tasks. 
The target tasks include depth estimation, semantic segmentation, and boundary detection, while self-supervised tasks include rotations, MoCo, and DenseCL.
We find that jointly optimizing with self-supervised auxiliary tasks consistently outperforms ImageNet-pretrained baselines.
The benefits of CompL are most pronounced in low-data regimes, where the importance of inductive biases increases~\cite{baxter2000model}. 
We also find that jointly optimizing monocular depth estimation with a self-supervised objective can outperform most labeled auxiliary tasks.
CompL can additionally improve semantic segmentation and boundary detection model robustness, when evaluated on new domains.
Our experiments demonstrate the promise of self-supervision as an auxiliary task. 
We envision these findings will establish the design of novel task-specific self-supervised methods as a new axis of investigation for future multi-task learning research.

%% file: text/relwork.tex
\section{Related Work}
\parsection{Multi-Task Learning (MTL)}
MTL aims to enhance performance and robustness of a predictor by jointly optimizing a shared representation between several tasks~\cite{caruana1997multitask}.
This is accomplished by exploiting the domain-specific information contained in the training signal of one task (e.g.,\ semantic segmentation), to more informatively select hypotheses for other tasks (e.g.,\ depth), and vice versa~\cite{ranjan2017hyperface,bruggemann2021exploring}. 
For example, pixels of class ``\emph{sky}'' will always have a larger depth that those of class ``\emph{car}''~\cite{saha2021learning}.
If non-related tasks are combined, however, the overall performance degrades.
This is referred to as task-interference and has been well documented in the literature~\cite{maninis2019attentive,kanakis2020reparameterizing}. 
However, no measurement of task relations can tell us whether performance gain can be achieved without training the final models.
Although several works have shown that while MTL can improve performance, it requires an exhaustive manual search of task interactions~\cite{standley2020tasks}, and labeled datasets with many tasks. 
In this work we also jointly optimize a network on multiple tasks, but we instead evaluate the efficacy of self-supervision as an auxiliary task. 
This enables the use of joint training in any dataset and eliminates expensive annotation efforts that do not guarantee performance gains.
To further improve performance of a target task, \cite{hoyer2021three,guizilini2020robust,baek2022semi,georgescu2021anomaly} designed specialised architectures for a predefined set of tasks. 
These architectures do not generalize to other tasks.
On the other end, \cite{liu2019self} aim to learn a sub-class labelling problem as an auxiliary task, i.e.\ for class dog learn the breed subclass, however the notion of subclass does not generalize to dense tasks like depth estimation.
Instead, we conduct a systematic investigation using a common pipeline, applicable to any dense target task.
This enables the easy switching of different supervised target tasks or auxiliary self-supervised tasks, without requiring any architectural changes, enabling the wider reach of joint training across tasks and datasets.

\parsection{Transfer learning} 
Given a large labeled dataset, neural networks can optimize for any task, whether image-level~\cite{krizhevsky2012imagenet}, or dense~\cite{he2019rethinking}. 
In practice, however, large datasets can be prohibitively expensive to acquire, giving rise to the transfer learning paradigm.
The most prominent example of transfer learning is the fine-tuning of an ImageNet~\cite{deng2009imagenet} pre-trained model on target tasks such as semantic segmentation~\cite{long2015fully}, or monocular depth estimation~\cite{fu2018deep}. 
However, ImageNet models do not always provide the best representations for all downstream tasks, raising interest in finding task relationships for better transfer capabilities~\cite{zamir2018taskonomy}. 
In this work we are not interested in learning better pre-trained networks for knowledge transfer.
Rather, we start from strong transfer learning baselines and improve generalization by jointly optimizing the target and auxiliary tasks.

\parsection{Self-supervised learning}
Learning representations that can effectively transfer to downstream tasks, coupled with the cost associated with the acquisition of large labeled datasets, has given rise to self-supervised methods. 
These methods can learn representations through explicit supervision on pre-text tasks~\cite{doersch2015unsupervised,gidaris2018unsupervised}, or through contrastive methods~\cite{chen2020simple,he2020momentum}. 
Commonly, self-supervised methods aim to optimize a given architecture, yielding better pre-training models for fine-tuning on the target task~\cite{doersch2015unsupervised,gidaris2018unsupervised,chen2020simple,he2020momentum,ghiasi2021multi,wang2020dense,newell2020useful,li2022univip}.
We instead utilized such pre-trained models as a starting point and fine-tune on both the target and self-superivsed auxiliary tasks jointly, rather than just the target task, to further improve performance and robustness.
More recently, supervised tasks have been used in conjunction with self-supervised techniques by exploiting the labels to guide contrastive learning.
This can be seen as a form of sampling guidance and has been utilized in classification~\cite{khosla2020supervised}, semantic segmentation~\cite{wang2021exploring}, and tracking~\cite{pang2021quasi}.
These methods differ from our work as they require target task labels to optimize the self-supervised objective, while our self-supervised objectives are independent of the target labels and can be applied on any set of images.
Instead, \cite{deng2021does} jointly train a model for classification and rotation, but utilize the rotation performance at test time as a proxy to the classification performance.
The goal of this work is instead to improve the target task's performance and robustness.
More closely to our work, \cite{gidaris2019boosting} and \cite{zhai2019s4l} jointly train classification and self-supervised objectives under a semi-supervised training protocol.
We also perform joint training with a self-supervised task, however, we follow a more general MTL methodology, and investigate whether self-supervised tasks can provide inductive bias to dense tasks.

\parsection{Robustness}
Robust predictors are important to ensure their performance under various conditions during deployment.
Recent works have focused on improving different aspects of robustness, such as image corruption~\cite{hendrycks2019benchmarking}, adversarial samples~\cite{zhang2019theoretically}, and domain shifts~\cite{yu2020bdd100k}.
More related to our work, \cite{hendrycks2019using} jointly train classification and self-supervised rotation, demonstrating that the strong regularization of the rotations improves model robustness to adversarial examples, and label or input corruptions. 
\cite{wang2021robust} similarly used joint training but employed both image and video-level self-supervised tasks and found them to improve the model's robustness to domain shifts for object detection.
We also evaluate the effect of joint training on robustness to unseen datasets, but focus on dense prediction tasks.

%% file: text/method.tex
\section{Composite Learning}

In this section, we introduce and motivate Composite Learning (CompL). 
Specifically, Sec.~\ref{method} formalizes the problem setting, Sec.~\ref{ss_methods} describes the self-supervised methods investigated, and Sec.~\ref{architecture} lists the network structure choices in our study.

\subsection{Joint Learning with Supervised and Self-Supervised Tasks}
\label{method}

Multi-task learning may improve the model robustness and generalizability. We aim to investigate the efficacy of joint training with self-supervision on dense prediction tasks as the targets.
The shared representation between the target task $t$ and an auxiliary task $a$ may be more effective than training on $t$ alone.

In the traditional MTL setup, the label sets $Y_t$, and $Y_a$, are manually labeled. 
In contrast, the auxiliary labels $Y_a$ in CompL are implicitly created in the self-supervised task.
Formally, CompL aims to produce the two predictive functions $f_t(\theta_{s},\theta_{t}): \mathcal{X}_t \rightarrow \mathcal{Y}_t$ and $f_a(\theta_{s},\theta_{a}): \mathcal{X}_a \rightarrow \mathcal{Y}_a$, where $f_t$ and $f_a$ share parameters $\theta_{s}$ and have disjoint parameters $\theta_{\{t,a\}}$.
During inference we are only interested in $f_t$, however, we hypothesize that we can learn a more effective parameterization through the above weight sharing scheme. 
In our investigation, $f_t$ and $f_a$ are trained jointly using samples $(X_t, y_t)$ and $(X_a, y_a)$.

The overall optimization objective therefore becomes
\begin{equation}
\vspace{-0.1in}
\min_{\theta_{s}, \theta_{t}, \theta_{a}} \mathcal{L}^{t}((X_t,y_t); \theta_{s}, \theta_{t})
+ \lambda \mathcal{L}^{a}((X_a,y_a); \theta_{s}, \theta_{a}),
\end{equation}

\noindent where $\mathcal{L}^{t}$ and $\mathcal{L}^{a}$ are the losses for the supervised and self-supervised tasks respectively, and $\lambda$ is a scaling factor controlling the magnitude and importance of the self-supervised task.

The experiments in this paper use the same dataset for both the target and auxiliary tasks. 
We additionally train our models using different-sized subsets $(X'_t, y'_t)$ for the target task, where $X'_t \subseteq X_t = X_a$.
However, the above is not a necessary condition for CompL, meaning the self-supervised task could be trained on an independent dataset.

\parsection{Training method}
We jointly optimize two objectives. 
We construct a minibatch by sampling at random independently from the two training sets. 
For simplicity, we sample an identical number of images from each training set. 
The input images $X_t$ and $X_a$ are treated independently.
This enables us to apply task/method-specific augmentations to each task input without causing task conflicts. We apply the baseline augmentations to $X_t$, ensuring a fair comparison with our single-task baselines. 
$X_a$ used for self-supervised training is instead processed with the proposed task-specific augmentations for each method investigated.
These augmentations include Gaussian blur and rotation. 
They can significantly degrade performance for dense tasks if applied on the target task, but they are important for self-supervision.
Therefore, by using distinct augmentations on two tasks, we can minimize performance degradation brought by training the auxiliary tasks.

\subsection{Self-Supervised Methods in Our Study}
\label{ss_methods}

\parsection{Rotation (Rot)} 
\cite{gidaris2018unsupervised} proposed to utilize 2-dimensional rotations on the input images to learn feature representations. 
Specifically, they optimize a classification model to predict the rotation angles, equally spaced in $[0^{\circ},360^{\circ} )$. 
Joint optimization with self-supervised rotation has demonstrated success in semi-supervised image classification~\cite{gidaris2019boosting,zhai2019s4l}, and enhanced robustness to input/output corruptions~\cite{hendrycks2019using}, making it a prime candidate for further investigation in a dense prediction setting.

\parsection{Global contrastive}
Global contrastive methods treat every image as its own class, while artificially creating novel instances of said class through random data augmentations. 
In this work, we evaluate contrastive methods using Momentum Contrast (MoCo)~\cite{he2020momentum}, and specifically MoCo v2~\cite{chen2020improved}. 
These methods formulate contrastive learning as dictionary look-up, enabling for the construction of a large and consistent dictionary of size $|Z|$ without the need for large batch sizes, a common challenge amongst dense prediction tasks~\cite{chen2018encoder}. 
MoCo is optimized using InfoNCE~\cite{oord2018representation}, a contrastive loss function defined as
\vspace{-0.05in}
\begin{equation}
\vspace{-0.05in}
\mathcal{L} = -\log \dfrac{\exp{(z^+/\tau)}}{\displaystyle\sum_{z\in Z} \exp{(z/\tau)}
}.
\end{equation}

\noindent
InfoNCE is a softmax-based classifier that optimizes for distinguishing the positive representation $z^+$ from the $|Z|-1$ negative representations. 
The temperature $\tau$ is used to control the smoothness of the probability distribution, with higher values resulting in softer distributions. 

\parsection{Local contrastive}
In dense predictions tasks, we desire a fine-grained pixel wise prediction rather than a global one. 
As such, we further investigate the difference between global contrastive MoCo v2~\cite{chen2020improved}, and its variant DenseCL~\cite{wang2020dense}, that includes an additional contrastive loss acting on local representations.

\begin{table*}[ht]
  \center
    \caption{Monocular depth estimation performance in RMSE on NYUD-v2. `$\rightarrow$' denote transfer learning methods, while `+' denote joint training (CompL). Initialization with DenseCL coupled with DenseCL joint training outperforms all other methods.}
    \vspace{-0.1in}
      \resizebox{\linewidth}{!}{%
      \footnotesize
  \begin{tabularx}{\linewidth}{@{}lcXXXXr@{}}
    \toprule
\multirow{2}{*}{Model} &&    \multicolumn{5}{c}{Labeled Data} \\
    \cmidrule{3-7} 
    && 5\% & 10\% & 20\% & 50\% & 100\% \\
    \midrule
	Depth && 0.8871  & 0.8120 & 0.7471  & 0.6655 & 0.6223 \\
	    \midrule
	Rot $\rightarrow$ Depth && 1.0830 & 1.0120 & 0.9114  & 0.8322 & 0.7822 \\
	MoCo $\rightarrow$ Depth && 0.8758 & 0.7708 & 0.7113 & 0.6311 & 0.5890 \\
	DenseCL $\rightarrow$ Depth && 0.8736 & 0.7726 & 0.7152 & 0.6321 & 0.5982  \\
	    \midrule
	Depth + Rot && 0.8762 & 0.8071 & 0.7298 & 0.6460 & 0.6107 \\
	Depth + MoCo && 0.8501 & 0.7955 & 0.7206 & 0.6434 & 0.6000 \\
	Depth + DenseCL && 0.8479 & 0.7866 & 0.7131 & 0.6420 & 0.5990 \\
	    \midrule
	MoCo $\rightarrow$ MoCo + Depth && 0.8614  & 0.7732 & 0.7008 & 0.6220 & 0.5773 \\
	DenseCL $\rightarrow$ DenseCL + Depth && \textbf{0.8468}  & \textbf{0.7641} & \textbf{0.6989} & \textbf{0.6157} & \textbf{0.5690} \\
    \bottomrule
  \end{tabularx}}
  \vspace{-0.1in}
\label{table:depth_joint_perf}
\end{table*}

\begin{figure*}[ht]
 \centering
    \includegraphics[width=0.9\textwidth]{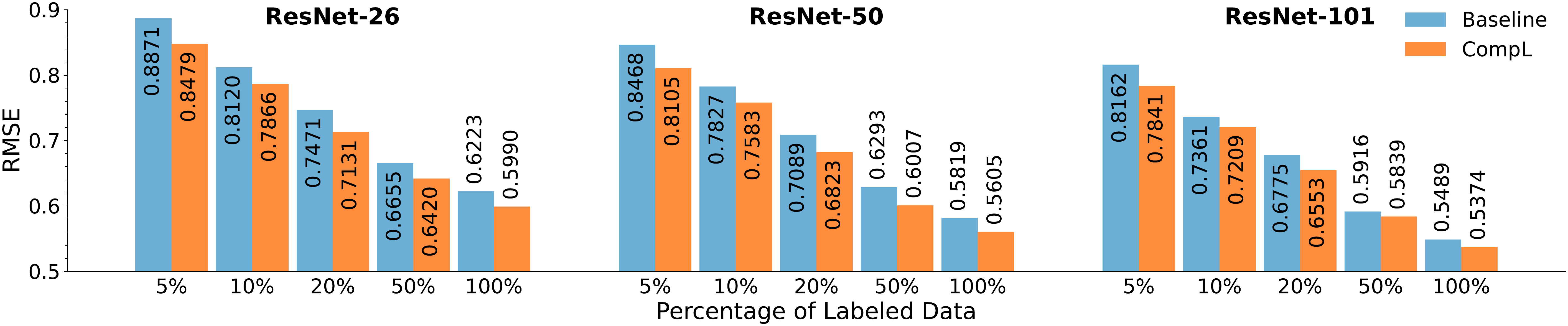}
\caption{Monocular depth estimation performance in RMSE on different ResNet encoders. Use of CompL (orange) denotes the addition of the best performing self-supervised objective (DenseCL). CompL consistently outperforms the baselines in all experiments.}
\label{fig:depth_encoders}
\vspace{-0.2in}
\end{figure*}

\subsection{Network Structures}
\label{architecture}

Dense prediction networks are initially pre-trained on classification, and then modified according to the downstream task of interest, e.g., by introducing dilations~\cite{yu2015multi}. 
In our investigation, we jointly optimize heterogeneous tasks such as a dense prediction task and image rotations. 
Therefore, our networks call for special structure considerations. 
This section presents the details.

\parsection{Dense prediction networks}
Common dense prediction networks use an encoder-decoder structure~\cite{ronneberger2015u,badrinarayanan2017segnet}, maintain a constant resolution past a certain network depth~\cite{yu2017dilated}, or even utilize both high and low representation resolutions in multiple layers of the network~\cite{wang2020deep}. 
Due to the large differences among networks, we opt to treat the entire network as a single unit, and only utilize the last feature representation of the networks for the task-specific predictions. 
In other words, we branch out at the last layer and employ a single task-specific module for the predictions.
This ensures that our findings do not depend on network structures, and it is easy to generalize to new network designs.

We perform our experiments on DeepLabv3+~\cite{chen2018encoder} based on ResNets~\cite{he2016deep}. The networks demonstrated competitive performance on a large number of dense prediction tasks, such as semantic segmentation, and depth estimation and has been used extensively when jointly learning multiple tasks~\cite{maninis2019attentive,bruggemann2020automated}. 
Our investigation is primarily on the smaller ResNet-26 architecture for easy comparison with existing MTL results.
As it is a common practice in dense prediction tasks, we initialize the ResNet encoder with ImageNet pre-trained weights, unless stated otherwise.

\parsection{Task-specific heads}
The final representation of the dense prediction networks is utilized in two task-specific modules. 
The first module, consisting of a 1\texttimes1 convolutional layer, generates the predictions of the supervised task, with the output dimension being task dependent, such as the number of classes. 
The second prediction head is specific for self-supervised tasks. 
Unlike the supervised prediction head, the self-supervised prediction head is utilized only during network optimization, and is discarded at test time.
The features for Rot and MoCo are first pooled with a global average pooling layer.
Rot is then processed by an fully connected layer with output dimensions equal to 4, number of potential rotations, while MoCo is processed by 2-layer MLP head with output dimensions equal to 128, feature embedding dimension.
DenseCL, on the other hand, generates two outputs. 
The first one is identical to MoCo for the global representation, while for the second representation, the initial dense features are pooled to a smaller grid size, and then processed with two 1\texttimes1 convolutional layer to get the local feature representations.

\begin{figure*}[ht]
\vspace{-0.1in}
 \centering
    \subfloat[Monocular depth estimation.\label{fig:depth_tsne}]{\includegraphics[width=0.35\textwidth]{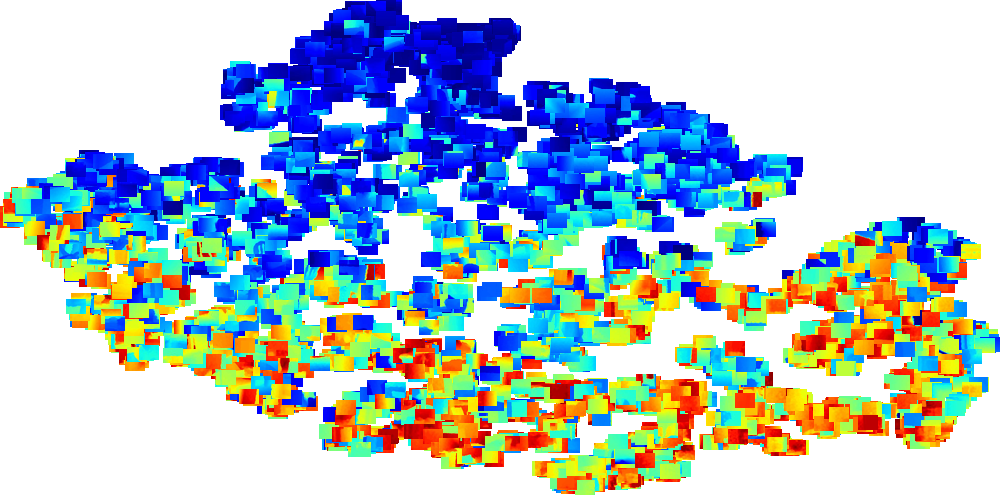}}~\qquad~\qquad
    \subfloat[Semantic segmentation.\label{fig:semsegdensecl_tsne}]{\includegraphics[width=0.35\textwidth]{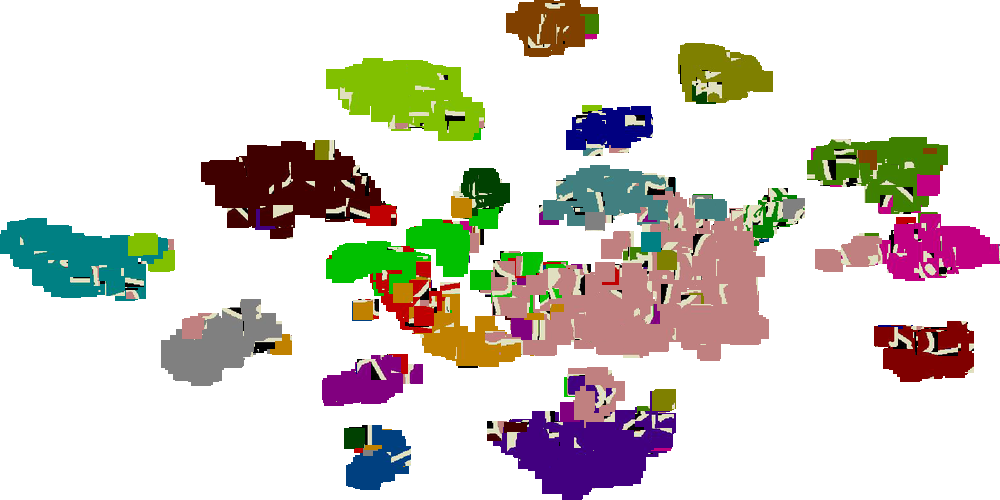}}
\caption{t-SNE visualization of the DenseCL local representations. 
The representations are depicted using their ground-truth maps. Specifically, (a) depth values for monocular depth estimation and (b) semantic patches for semantic segmentation.
The local representations adapt to the target task, i.e., (a) smooth depth variation for the regression task while (b) clusters are formed for the classification task.} \label{fig:densecl_tsne}
\vspace{-0.1in}
\end{figure*}

\begin{figure}[ht]
 \centering
    \includegraphics[width=0.45\textwidth]{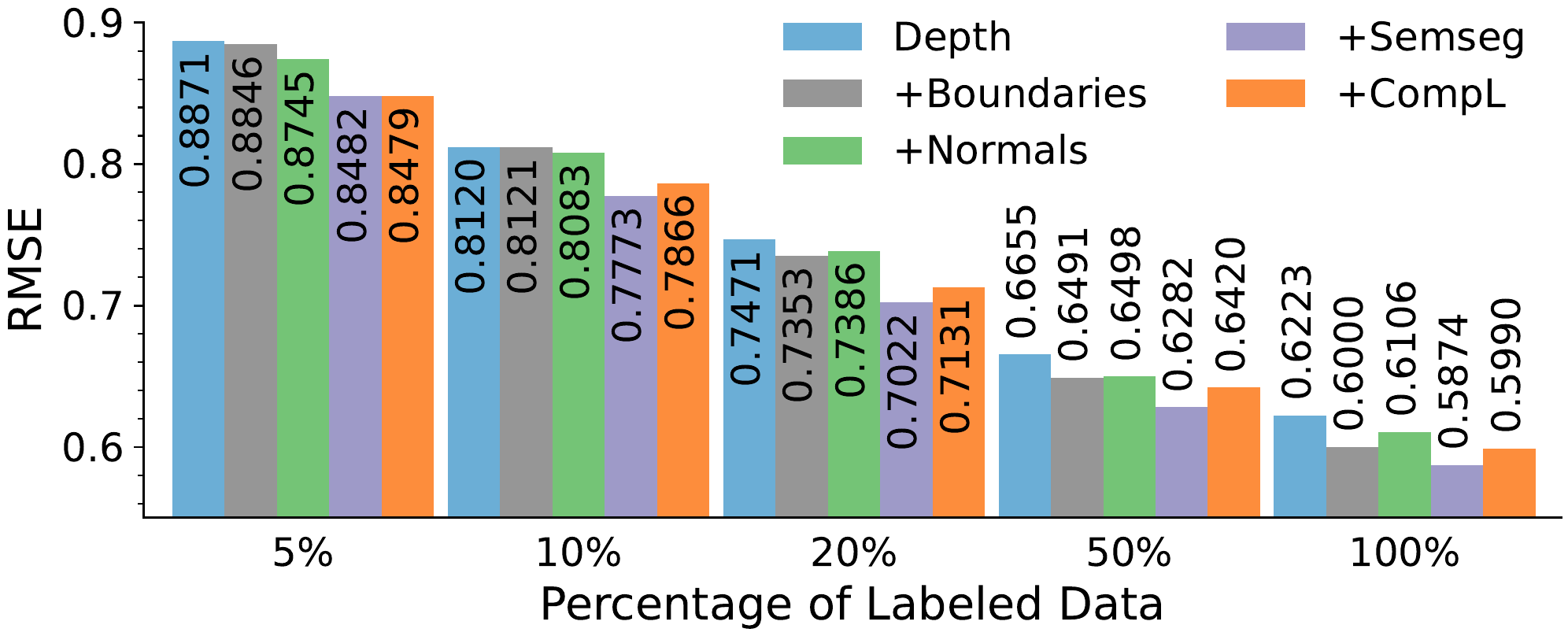}
\caption{Monocular depth estimation performance in RMSE on NYUD-v2 when trained with additional auxiliary tasks. CompL can improve depth more than training with boundary or normal predictions. Semantic segmentation can improve the depth prediction more, but it requires expensive manual annotations.}
\label{fig:depth_same_splits}
\vspace{-0.5in}
\end{figure}

\parsection{Normalization}
Large CNNs are often challenging to train, and thus utilize Batch Normalization (BN) to accelerate training~\cite{ioffe2015batch}. 
In self-supervised training, BNs often degrade performance due to intra-batch knowledge transfer among samples.
Workarounds include shuffling BNs~\cite{he2020momentum,chen2020improved}, using significantly larger batch sizes~\cite{chen2020simple}, or even replacing BNs altogether~\cite{henaff2020data}.
To ensure BNs will not affect our study, and findings can be attributed to the jointly trained tasks, we replace BNs with group normalization (GN)~\cite{wu2018group}. 
We chose GN as it yielded the best performance when trained on ImageNet~\cite{wu2018group}. However, other normalization layers that are not affected by batch statistics can also be utilized, such as layer~\cite{ba2016layer} and instance~\cite{ulyanov2016instance} normalizations.

%% file: text/exp.tex
\begin{table*}[ht]
  \center
    \caption{Semantic segmentation performance in mIoU on the PASCAL VOC dataset. `$\rightarrow$' denote transfer learning methods, while `+' denote joint training (CompL). Joint training with DenseCL significantly outperforms the ``Semseg'' baselines.}
    \vspace{-0.1in}
      \resizebox{\linewidth}{!}{%
      \footnotesize
  \begin{tabularx}{\linewidth}{@{}lcXXXXXXr@{}}
    \toprule
\multirow{2}{*}{Model} &&    \multicolumn{7}{c}{Labeled Data} \\
    \cmidrule{3-9} 
     && 1\% & 2\% & 5\% & 10\% & 20\% & 50\% & 100\% \\
    \midrule
	Semseg && 30.82  & 37.66 & 49.95  & 55.17 & 61.30  & 67.38 & 70.42  \\
	
	    \midrule
	Rot $\rightarrow$ Semseg && 10.35  & 12.43 & 18.29  & 24.71 & 29.21  & 35.43 & 39.46  \\
	MoCo $\rightarrow$ Semseg && 31.55 & 37.55 & 48.60  & 53.27 & 58.74  & 64.04 & 68.09  \\
	DenseCL $\rightarrow$ Semseg && 34.89 & 39.72 & 50.96  & 55.60 & 61.13  & 65.71 & 69.56  \\
	    \midrule
	Semseg + Rot && 28.75 & 36.81 & 50.46 & 56.21 & 62.17 & 67.96 & 70.52 \\

	Semseg + MoCo && 32.90 & 40.31 & 52.18  & 56.50 & 62.49  & 68.40 & 71.15  \\
	
	Semseg + DenseCL && 33.51 & 40.91 & 52.76 & \textbf{57.33} & \textbf{63.22} & \textbf{68.81} & \textbf{71.16} \\
	\midrule
	DenseCL $\rightarrow$ Semseg + DenseCL && \textbf{36.32} & \textbf{41.24} & \textbf{52.94} & 56.87 & 62.71 & 65.89 & 69.81 \\
    \bottomrule
  \end{tabularx}}
\label{table:semseg_joint_perf}
\vspace{-0.1in}
\end{table*}

\begin{figure*}[ht]
\small
 \centering
    \includegraphics[width=0.9\textwidth]{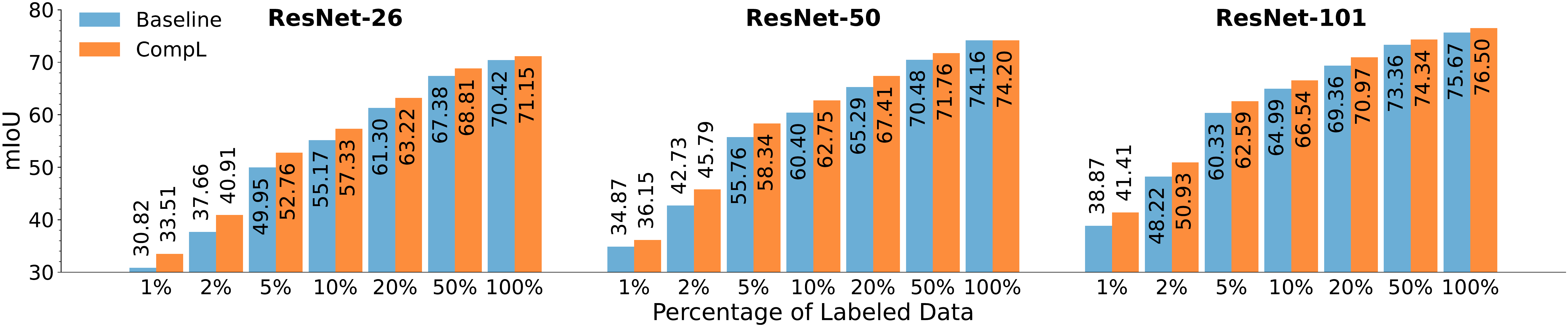}
\caption{Semantic segmentation performance in mIoU on different ResNet encoders. Use of CompL (orange) denotes the addition of the best performing self-supervised objective (DenseCL). CompL consistently outperforms the baselines in all experiments.}
\label{fig:semseg_encoders}
\vspace{-0.2in}
\end{figure*}

\section{Experiments}
\label{sec:exp}

In this section we investigate the effects of jointly training dense predictions and self-supervised tasks. 
To systematically assess the effect of joint learning in label-deficient cases, we use different-sized subsets $(X'_T, y'_t)$ of the full target task data $(X_T, y_t)$, i.e., $(X'_T, y'_t) \subseteq (X_T, y_t)$. 
To ensure consistent contribution from the auxiliary task, we always use the full data split $(X_A, y_a)$ for the self-supervised task. 
The supplementary material includes additional experiments using the same subsets for both tasks.

\parsection{Implementation details}
We sample 8 images at random from each of the target and auxiliary training sets. 
We apply the baseline augmentations to target samples, namely, random horizontal flipping, random image scaling in the range [0.5, 2.0] in 0.25 increments, and then crop or pad the image to ensure a consistent size. 
The auxiliary loss is scaled by $\lambda$.
We found 0.2 works best for MoCo and DenseCL, while 0.05 for Rot.
The model is optimized using stochastic gradient decent with momentum 0.9, weight decay 0.0001, and the ``poly'' learning rate schedule~\cite{chen2017deeplab}. 


\subsection{Monocular Depth Estimation}
\label{sec:mde}
We first evaluate CompL on monocular depth estimation. 
Monocular depth estimation is a widely used dense prediction task, and is typically casted as a regression problem.

\parsection{Experimental protocol}
Monocular depth estimation is explored on NYUD-v2~\cite{silberman2012indoor}, comprised of 795 train and 654 test images from indoor scenes, and evaluated using the root mean squared error (RMSE) metric. 
All models are trained for 20k iterations, corresponding to 200 epochs of the fully labeled dataset, with an input image size of 425\texttimes560, and are optimized with the $\mathcal{L}_1$ loss. 

\parsection{Joint optimization}
Table~\ref{table:depth_joint_perf} presents the performance of the single-task baseline, ``Depth'', and the models trained jointly with different self-supervised tasks, ``Depth + \emph{Task name}''. 
We find that joint training with any self-supervised task consistently improves the performance of the target task, even in the fully labeled dataset. 
In particular, joint training with self-supervision yields the biggest performance improvements on the lower labeled percentages, where the importance of inductive bias increases~\cite{baxter2000model}.
These findings are consistent also when utilizing stronger ResNet encoders, as depicted in Fig.~\ref{fig:depth_encoders} for the best performing self-supervised DenseCL method.

DenseCL contrasts both local and global representations, yielding richer representations for dense task pre-training, as compared to the image-level self-supervised tasks.
We find this to also be the case in our joint-training setup, where richer local representations help guide the optimization of depth.
To better understand the benefit of utilizing DenseCL for joint training with depth, we visualize the representations in Fig.~\ref{fig:depth_tsne} using a t-SNE plot~\cite{van2008visualizing}.
Specifically, we depict the latent representations of DenseCL using their corresponding ground-truth depth measurements.
The depth values smoothly transition from larger distances (in red) to smaller distances (in blue). 
This indicates that the DenseCL objective, which is discriminative by construction, promotes a smooth variation in the representations when combined with a regression target objective.

\parsection{Traditional MTL}
In order to determine how CompL compares to traditional MTL, we evaluate and compare the effect of using labeled auxiliary tasks.
Specifically, we investigate the effect of the remaining three tasks of NYUD-v2, that is, boundaries, normals, and semantic segmentation, in Fig.~\ref{fig:depth_same_splits}. 
For fair comparisons to CompL, the auxiliary tasks also use the entire dataset. 
CompL consistently outperforms the use of labeled boundaries and normals as auxiliary tasks.
This is particularly pronounced in the lower data splits where the contribution of CompL becomes more prominent, while boundaries and normals contribute less.
Surface normals, derivatives of depth maps, could be expected to boost depth prediction due to their close relationship.
However, we find it to help only marginally.
On the other hand, joint training with semantic segmentation consistently improves the baseline performance, which aligns with findings in the previous works~\cite{chen2019towards,guizilini2020semantically,jiao2018look}. 
These results exemplify the importance of an arduous iteration process in search of a synergistic auxiliary task, where knowledge of label interactions are not necessarily helpful.
This process is further complicated when additional auxiliary task annotations are needed.
Therefore, eliminating manual labeling from auxiliary tasks opens up a new axis of investigation for the future of multi-task learning research as it can enable faster iterations in task interaction research.

\parsection{Transfer learning}
The experiments have so far shown that joint training with self-supervision can enhance performance, and in most cases outperforms traditional MTL practices.
Notably, outperforming the baselines even when all models are initialized with ImageNet pre-trained weights, a strong transfer learning baseline.
However, is ImageNet pre-training the best initialization for Depth, and how does it compare to self-supervised pre-training?
In Table~\ref{table:depth_joint_perf} we repeat the baseline experiments starting from self-supervised pre-training, (``\emph{Initial task} $\rightarrow$ Depth'').
In depth estimation, the contrastive methods gain the advantage and outperform the joint training methods.
However, our proposed method is not limited by the initialization used.
We find that initialization with MoCo or DenseCL weights coupled with joint training (``\emph{Initial task} $\rightarrow$ \emph{Initial task} + Depth'') can increase the performance even further, giving the best performing models.


\subsection{Semantic Segmentation}
\label{sec:semseg_exp}

We additionally evaluate semantic segmentation. 
Semantic segmentation is representative for discrete labeling dense predictions.

\parsection{Experimental protocol}
Semantic segmentation (Semseg) experiments are conducted on PASCAL VOC 2012~\cite{everingham2010pascal}, and specifically the augmented version (aug.) from \cite{hariharan2011semantic}, that provides 10,582 train and 1,449 test images. 
We evaluate performance in terms of mean Intersection-over-Union (mIoU) across the classes. 
All models are trained for 80k iterations, accounting for 60 epochs of the fully labeled dataset, and are optimized with the cross-entropy loss with image input size of 512\texttimes512. 

\parsection{Joint optimization}
Table~\ref{table:semseg_joint_perf} present the performance of the single-task baseline and the models trained jointly with different self-supervised tasks. 
In contrast to findings from classification literature~\cite{gidaris2019boosting,zhai2019s4l}, joint training with Rot minimally affects the performance in most cases, with lower labeled percentages even incurring a performance degradation. 
On the other hand, the contrastive methods increase performance on all labeled splits, with lower labeled percentages incurring the biggest performance improvement. 
These findings are once again consistent when utilizing stronger ResNet encoders, as depicted in Fig.~\ref{fig:semseg_encoders} for the best performing self-supervised method DenseCL.
Similar to depth, we further visualize in Fig.~\ref{fig:semsegdensecl_tsne} the latent representations contrasted by DenseCL, and depict them with their ground-truth semantic maps. 
Unlike in depth regression, where the representations were smooth due to the continuous nature of the problem, the DenseCL representations for semantic segmentation form clusters given the discriminative nature of semantic segmentation.

\begin{figure}[t]
 \centering
  \includegraphics[width=0.35\textwidth]{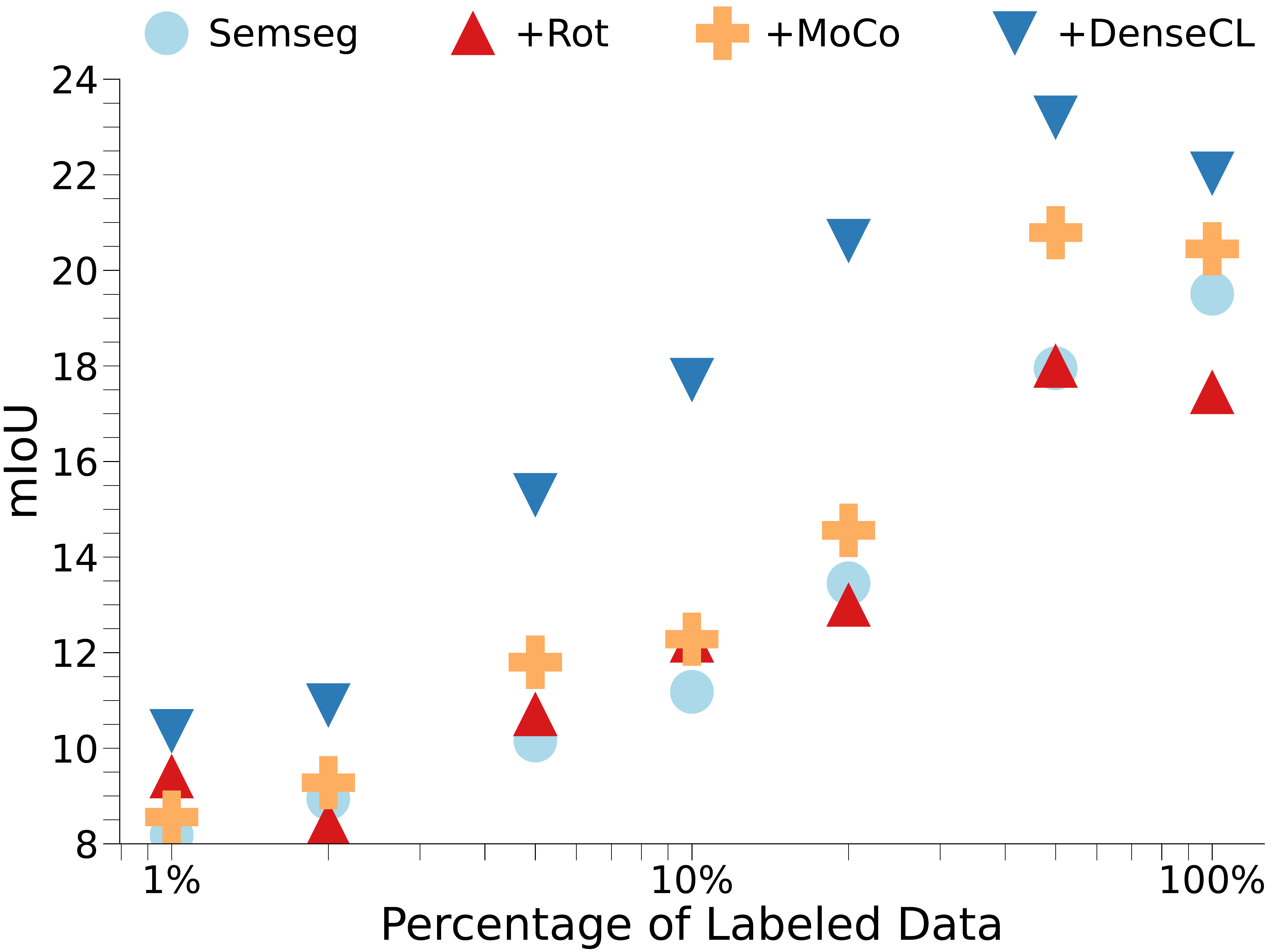}
\caption{Semantic segmentation performance in mIoU trained on PASCAL VOC and evaluated on BDD100K. The local contrastive loss of DenseCL provides significant robustness improvements.}
\label{fig:semseg_zeroshot}
\vspace{-0.2in}
\end{figure}

\parsection{Robustness to zero-shot dataset transfer}
So far we have only evaluated on the same distribution as that used for training, however, distribution shifts during deployment are common.
We therefore investigate the generalization capabilities to new and unseen datasets. 
We evaluate the zero-shot capabilities of the models on the challenging BDD100K~\cite{yu2020bdd100k} dataset in Fig.~\ref{fig:semseg_zeroshot}, a diverse driving dataset. 
The test frames from BDD100K are therefore significantly different to those observed during training, making zero-shot transfer particularly interesting due to the large domain shift. 
We report the mIoU with respect to the shared classes between the two datasets. Please refer to the supplementary for the table of the BDD100K experiments.

We find that Rot often performs worse than the baseline model. 
This yields dissimilar findings to classification~\cite{hendrycks2019using} that observed increased robustness, attributed to the strong regularization induced by the joint training.
For Semseg, such regularizations degrade the fine-grained precision required. 
Joint training with DenseCL significantly outperforms all other self-supervised methods. 
While MoCo was comparable to DenseCL on VOC (Table~\ref{table:semseg_joint_perf}), we find that local contrastive plays a big role in improving robustness. 
Interestingly, when using 100\% of the data points, performance on all methods utilizing self-supervision is lower than when using 50\% of the labels. 
We conjecture that, using the fully labeled split decreases the influence of self-supervision, making the model more prone to overfit to the training dataset and loose generalizability. 

\begin{table}[t!]
  \center
  \caption{Boundary detection performance in ODS F-score on the BSDS500 dataset. `$\rightarrow$' denote transfer learning methods, while `+' denotes joint training. Performance improvements are marginal, in contrast to the findings for other target tasks.}
  \vspace{-0.1in}
      \resizebox{\linewidth}{!}{%
      \footnotesize
  \begin{tabularx}{\linewidth}{@{}lcXXXr@{}}
    \toprule
\multirow{2}{*}{Model} &&    \multicolumn{4}{c}{Labeled Data} \\
    \cmidrule{3-6} 
    && 10\% & 20\% & 50\% & 100\% \\
    \midrule
	Boundaries  && 71.10 & 73.50 & 75.90 & 76.80 \\
	\midrule
	Rot $\rightarrow$ Boundaries && 60.20 & 62.80 & 66.00 & 67.70 \\
	MoCo $\rightarrow$ Boundaries && 71.00 & 73.40 & 75.60 & 76.40 \\
	DenseCL $\rightarrow$ Boundaries && 68.90 & 71.70  & 75.40 & 75.90 \\
    \midrule
	Boundaries + Semseg && 70.60 & 73.30 & 75.60 & \textbf{76.90} \\
	\midrule
	Boundaries + Rot  && 69.70 & 73.00 & 75.70 & 76.60 \\
	Boundaries + MoCo && \textbf{71.30} & 73.80 & \textbf{76.20} & \textbf{76.90} \\
	Boundaries + DenseCL && \textbf{71.30} & \textbf{73.90} & 76.00 & 76.20 \\
    \bottomrule
  \end{tabularx}}
\label{table:edge_joint_perf}
\vspace{-0.2in}
\end{table}

\parsection{Transfer learning}
Table~\ref{table:semseg_joint_perf} additionally reports the baseline experiments starting from self-supervised pre-training (indicated by ``\emph{Initial task} $\rightarrow$ Semseg''), or additionally optimized with the best performing DenseCL method, as in the Depth experiments.
Joint training with self-supervision consistently outperforms the sequential training counterpart, and in the majority of the cases by a significant margin.
In other words, CompL consistently reports performance gains when initializing with either ImageNet or DenseCL.


\subsection{Boundary Detection}

Boundary detection is another common dense prediction tasks. 
Unlike depth prediction and semantic segmentation, the target boundary pixels only account for a small percentage of the overall pixels. 
We find that CompL significantly improves the model robustness for boundary detection.

\parsection{Experimental protocol}
We study boundary detection on the BSDS500~\cite{arbelaez2010contour} dataset, consisting of 300 train and 200 test images. Since the ground truth labels of BSDS500 are provided by multiple annotators, we follow the approach of~\cite{xie2015holistically} and only count a pixel as positive if it was annotated as positive by at least three annotators. Performance is evaluated using the Optimal-Dataset-Scale F-measure (ODS F-score)~\cite{martin2004learning}. All models are trained for 10k iterations on input images of size 481\texttimes481. Following \cite{xie2015holistically}, we use a cross-entropy loss with a weight of 0.95 for the positive and 0.05 for the negative pixels.

\begin{table*}[t]
  \center
    \caption{Performance of a multi-task model for monocular depth estimation in RMSE and semantic segmentation in mIoU on NYUD-v2. `+' denote joint training. The multi-task model combined with CompL yields consistent improvements in both tasks.}
\vspace{-0.1in}
      \resizebox{\linewidth}{!}{%
      \footnotesize
  \begin{tabularx}{\linewidth}{@{}lXXXXXcXXXXr@{}}
    \toprule
 \multirow{2}{*}{Model} &    \multicolumn{5}{c}{Depth Labeled Data $\downarrow$} &&    \multicolumn{5}{c}{Semseg Labeled Data $\uparrow$} \\
    \cmidrule{2-6} \cmidrule{8-12} 
    & 5\% & 10\% & 20\% & 50\% & 100\% && 5\% & 10\% & 20\% & 50\% & 100\% \\
    \midrule
	Depth + Semseg & 0.997 & 0.904 & 0.794 & 0.665 & 0.606 && 10.46 & 14.99 & 19.41 & 26.24 & 31.66 \\
	    \midrule
	Depth + Semseg + DenseCL & \textbf{0.902} & \textbf{0.806} & \textbf{0.744} & \textbf{0.641} & \textbf{0.590} && \textbf{10.72} & \textbf{15.29} & \textbf{20.08} & \textbf{28.18} & \textbf{33.48} \\
    \bottomrule
  \end{tabularx}}
  \label{table_sup:depth_semseg_joint_perf}
  \vspace{-0.2in}
\end{table*}

\parsection{Joint optimization}
Table~\ref{table:edge_joint_perf} presents the performance of the single-task baseline and the models trained jointly with different self-supervised tasks. 
Compared to the previous two tasks, boundary detection is marginally improved by CompL.
Since convolutional networks are biased towards recognising texture rather than shape~\cite{geirhos2018imagenet}, we hypothesize that the supervisory signal of contrastive learning interferes with the learning of edge / shape filters essential for boundary detection. 
To investigate this hypothesis further, we jointly train boundary detection with a labeled high-level semantic task. 
Specifically, we jointly train boundary detection with the ground-truth foreground-background segmentation maps for BSDS500~\cite{arbelaez2010contour} from~\cite{endres2010category}. 
As seen in Table~\ref{table:edge_joint_perf}, the incorporation of semantic information once again does not enhance the single-task performance of boundaries, and even slightly degrades at lower percentage splits. 

While CompL yielded performance improvements for monocular depth estimation and semantic segmentation as target tasks, boundary estimation does not observe the same benefits. 
This further demonstrates the complexity of identifying a universal auxiliary task for all target tasks.
Instead, it demonstrates the importance of co-designed self-supervised tasks alongside the downstream task.

\begin{figure}[t]
 \centering
  \includegraphics[width=0.35\textwidth]{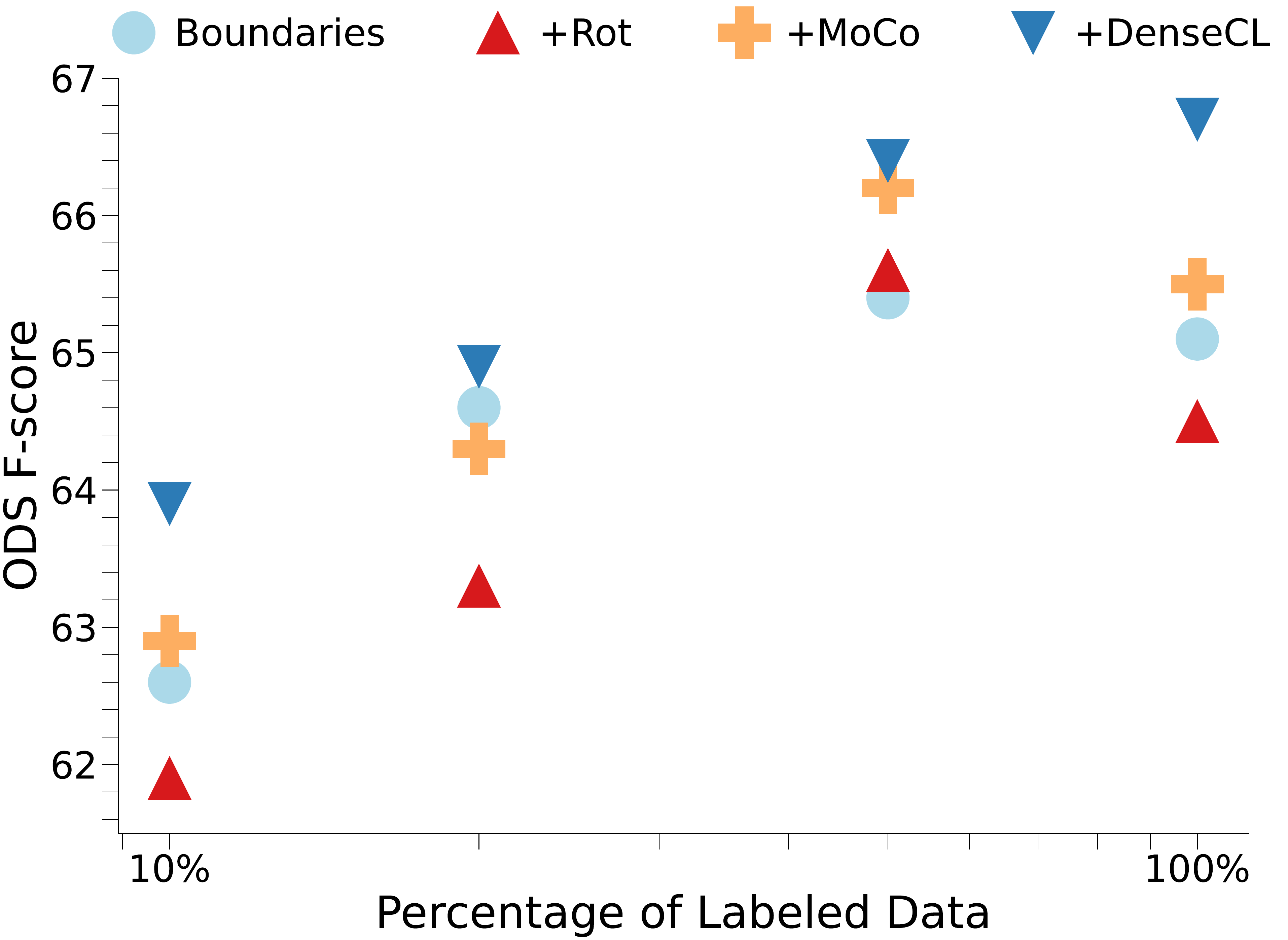}
\caption{Boundary detection performance in ODS F-score trained on BSDS and evaluated on NYUD. The additional local contrast of DenseCL increases robustness to zero-shot dataset transfer.}
\label{fig:sup_boundary_zeroshot}
\vspace{-0.2in}
\end{figure}

\parsection{Robustness to zero-shot dataset transfer}
We evaluate the zero-shot dataset transfer capabilities of the BSDS500~\cite{arbelaez2010contour} models from Table~\ref{table:edge_joint_perf} on NYUD-v2~\cite{silberman2012indoor}. 
Interestingly, even though CompL did not significantly improve the performance in Table~\ref{table:edge_joint_perf}, we find that the robustness experiments in Fig.~\ref{fig:sup_boundary_zeroshot} paint a different picture.
While MoCo often outperformed DenseCL in Table~\ref{table:edge_joint_perf}, and most methods perform comparatively to the baseline, the additional local constrast of DenseCL significantly improves the robustness experiments. 
This can be seen from DenseCL consistently outperforming the baseline, as well as all other methods.

\parsection{Transfer learning}
Table~\ref{table:edge_joint_perf} also reports the performance of the boundary detection transfer learning experiments. All three transfer learning approaches fare worse than ImageNet initialization, corroborating our hypothesis that boundary detection requires representations which are fairly unrelated to the features learned through self-supervision.


\subsection{Multi-Task Model (Semseg and Depth)}

Both semantic segmentation (Semseg) and monocular depth estimation (Depth) observed improvements when trained under CompL. In this section, we further investigate the applicability of CompL on MTL models optimized jointly for Depth and Semseg (Depth + Semseg).

\parsection{Experimental protocol}
We explore joint training on NYUD-v2~\cite{silberman2012indoor}, which provides ground-truth labels for both tasks. 
We maintain the exact same hyperparameters as the models in Sec.~\ref{sec:mde}, however, we expect an explicit search could yield additional improvements. 
No additional task-specific scaling of the losses is used, following \cite{maninis2019attentive}. 
For self-supervised tasks, we only evaluate DenseCL~\cite{wang2020dense}, as it performed the best for both tasks independently.

\parsection{Joint optimization}
Table~\ref{table_sup:depth_semseg_joint_perf} presents the performance of the baseline multi-task model (Depth + Semseg) and the model trained jointly with DenseCL (Depth + Semseg + DenseCL). 
As in the single-task settings, training under CompL enhances the performance of both Semseg and Depth. 
Specifically, we again observe a performance gain in every labeled percentage. 
This demonstrates that, even in the traditional multi-task setting, the additional use of CompL has the potential of yielding further performance gains. 
In the current setting, Depth observes a noticeable gain over Semseg in low data regimes. 
This can be attributed to the DenseCL hyperparameters being optimized directly for the improvement of Depth.
More advanced loss balancing schemes~\cite{chen2018gradnorm} could yield a redistribution of the performance gains, however, such investigation is beyond the scope of our work.

%% file: text/conclusion.tex
\section{Conclusion}

In this paper we introduced CompL, a method that exploits the inductive bias provided by a self-supervised task to enhance the performance of a target task. 
CompL exploits the label-free supervision of self-supervised methods, facilitating faster iterations through different task combinations.
We show consistent performance improvements in fully and partially labeled datasets for both semantic segmentation and monocular depth estimation.
While our method eliminated the need for labeling the auxiliary task, it commonly outperforms the traditional MTL with labeled auxiliary tasks on monocular depth estimation.
Additionally, the semantic segmentation models trained under CompL yield better robustness on zero-shot cross dataset transfer. 
We envision our contribution to spark interest in the explicit design of self-supervised tasks for their use in joint training, opening up a new axis of investigation for future multi-task learning research.

%% file: text/additional_depth.tex
\section{Monocular Depth Estimation}

\parsection{Joint optimization on identical dataset subsets}
Table~\ref{table:sup_depth_same_splits} presents the performance of the monocular depth estimation single-task baseline and the best performing self-supervised task, DenseCL. While in the main paper 
(Sec.~\ref{sec:mde}, Table~\ref{table:depth_joint_perf}) 
the self-supervised objective had access to the entire dataset, in Table~\ref{table:sup_depth_same_splits} both objectives use the same subset for optimization. Consistent improvements across all dataset splits are still observed.

\begin{table}[ht]
  \center
    \caption{Monocular depth estimation performance in RMSE on NYUD-v2.
  Both supervised and self-supervised objectives use identical splits. CompL denotes the addition of the best performing self-supervised objective, DenseCL, and yields consistent improvements.}
  \vspace{-0.1in}
      \resizebox{\linewidth}{!}{%
      \footnotesize
  \begin{tabularx}{\linewidth}{@{}ccXXXXr@{}}
    \toprule
\multirow{2}{*}{CompL}&&  \multicolumn{5}{c}{Dataset Size} \\
    \cmidrule{3-7} 
     && 5\% & 10\% & 20\% & 50\% & 100\% \\
    \midrule
	&& 0.8871  & 0.8120 & 0.7471  & 0.6655 & 0.6223 \\
	\checkmark&& \textbf{0.8840} & \textbf{0.8080} & \textbf{0.7305} & \textbf{0.6508} & \textbf{0.5990} \\
    \bottomrule
  \end{tabularx}}
  \label{table:sup_depth_same_splits}
\end{table}

%% file: text/additional_semseg.tex
\section{Semantic Segmentation}

\parsection{Joint optimization on identical dataset subsets}
Table~\ref{table:sup_semseg_same_splits} presents the performance of the semantic segmentation single-task baseline and the best performing self-supervised task DenseCL. Similar to Table~\ref{table:sup_depth_same_splits},
both objectives use the same subset for optimization. Consistent improvements across all dataset splits are again observed.

\parsection{Robustness to zero-shot dataset transfer}
In 
Sec.~\ref{sec:semseg_exp}
, we additionally investigated the generalization capabilities of CompL to a new and unseen dataset.
Table~\ref{table:sup_bdd_semseg_zeroshot} presents the performance of the BDD100K experiments from 
Fig.~\ref{fig:semseg_zeroshot}.

We additionally evaluate how the models trained on PASCAL VOC from 
Table~\ref{table:semseg_joint_perf}
(``Semseg'' and ``Semseg + \emph{Task name}'') perform without re-training on COCO~\cite{lin2014microsoft} on the same classes. 
As seen in Table~\ref{table:sup_coco_semseg_zeroshot} and Fig.~\ref{fig:sup_semseg_coco_zeroshot}, joint training with the contrastive methods consistently outperform across all percentage splits, with the lower labeled percentages observing the biggest improvement.

%% file: text/mtl.tex
\section{Multi-Task Model (Semseg and Depth)}

\begin{table}[t]
  \center  
  \caption{Semantic segmentation performance in mIoU on PASCAL VOC.
  Both supervised and self-supervised objectives use identical splits. CompL denotes the addition of the best performing self-supervised objective, DenseCL, and yields consistent improvements.}
  \vspace{-0.1in}
      \resizebox{\linewidth}{!}{%
      \footnotesize
  \begin{tabularx}{\linewidth}{@{}ccXXXXXXr@{}}
    \toprule
\multirow{2}{*}{CompL}&&  \multicolumn{7}{c}{Dataset Size} \\
    \cmidrule{3-9} 
     && 1\% & 2\% & 5\% & 10\% & 20\% & 50\% & 100\% \\
    \midrule
	&& 30.82  & 37.66 & 49.95  & 55.17 & 61.30  & 67.38 & 70.42  \\
	\checkmark&& \textbf{31.59} & \textbf{38.85} & \textbf{50.87} & \textbf{56.45} & \textbf{61.92} & \textbf{68.06} & \textbf{71.15} \\
    \bottomrule
  \end{tabularx}}
  \label{table:sup_semseg_same_splits}
\end{table}

\begin{figure}[t]
 \centering
  \includegraphics[width=0.8\linewidth]{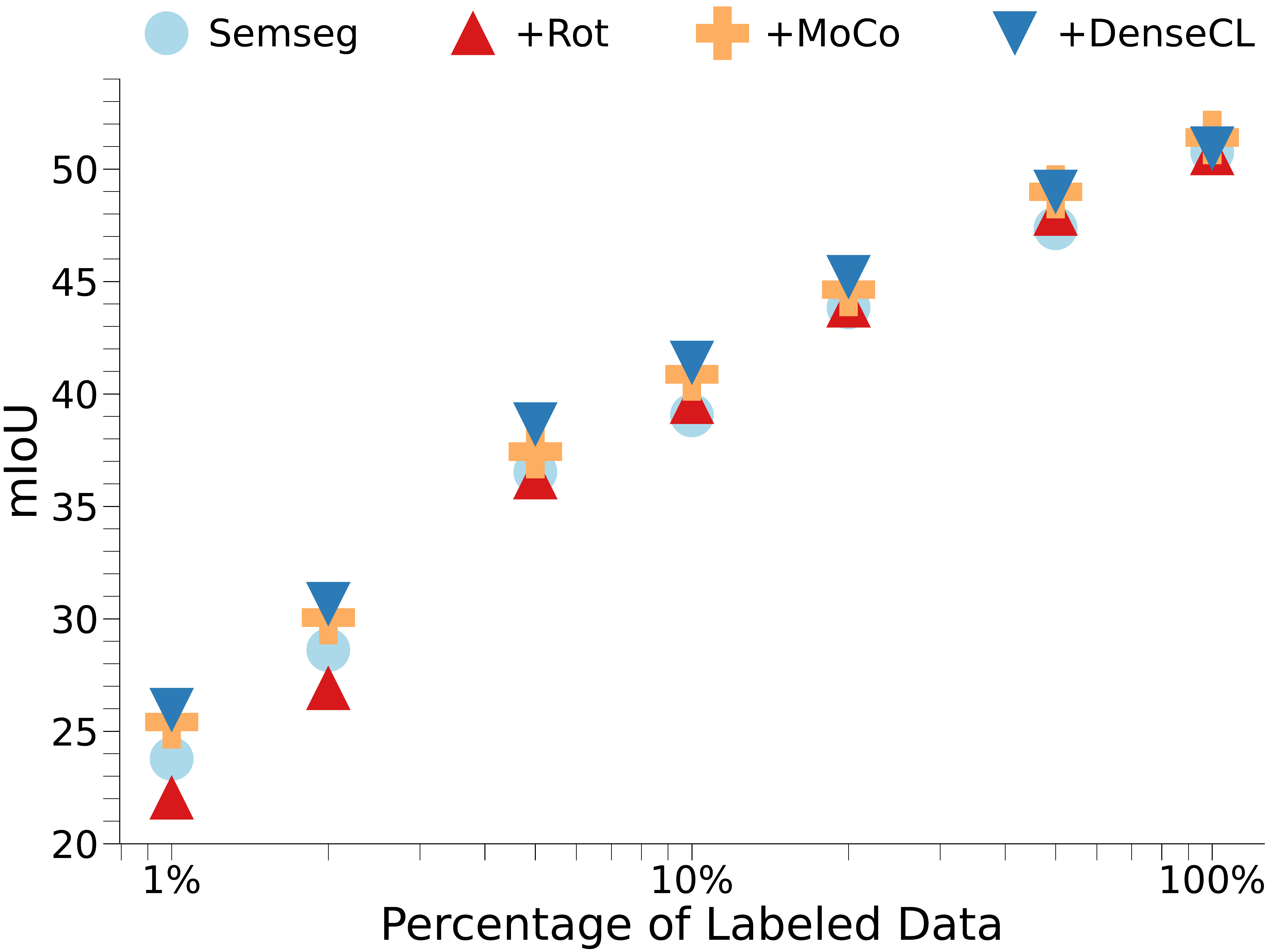}
\caption{Performance of semantic segmentation in mIoU trained on PASCAL VOC and evaluated on COCO. The local contrastive
loss of DenseCL provides consistent robustness improvements.}
\label{fig:sup_semseg_coco_zeroshot}
\end{figure}

\begin{table*}[t]
  \center
  \caption{Performance of semantic segmentation in mIoU trained on PASCAL VOC and evaluated on BDD100K. The local contrastive
loss of DenseCL provides significant robustness improvements.}
  \vspace{-0.1in}
      \resizebox{\linewidth}{!}{%
      \footnotesize
  \begin{tabularx}{\linewidth}{@{}lcXXXXXXr@{}}
    \toprule
\multirow{2}{*}{Model} &&    \multicolumn{7}{c}{Labeled Data} \\
    \cmidrule{3-9} 
     && 1\% & 2\% & 5\% & 10\% & 20\% & 50\% & 100\% \\
    \midrule
	Semseg &&  8.18 & 8.95& 10.16& 11.18& 13.45& 17.95& 19.51 \\
	    \midrule
	Semseg + Rot && 9.41 & 8.42 & 10.71 & 12.25 & 13.00 & 18.00 & 17.45\\
	Semseg + MoCo && 8.56 & 9.28 & 11.8 & 12.28 & 14.56 & 20.79 & 20.45\\
	Semseg + DenseCL && \textbf{10.36} & \textbf{10.90} & \textbf{15.30} & \textbf{17.71} & \textbf{20.62} & \textbf{23.20} & \textbf{22.03}\\
    \bottomrule
  \end{tabularx}}

\label{table:sup_bdd_semseg_zeroshot}
\end{table*}

\begin{table*}[t]
  \center
  \caption{Performance of semantic segmentation in mIoU trained on PASCAL VOC and evaluated on COCO. The local contrastive
loss of DenseCL provides consistent robustness improvements.}
  \vspace{-0.1in}
      \resizebox{\linewidth}{!}{%
      \footnotesize
  \begin{tabularx}{\linewidth}{@{}lcXXXXXXr@{}}
    \toprule
\multirow{2}{*}{Model} &&    \multicolumn{7}{c}{Labeled Data} \\
    \cmidrule{3-9} 
     && 1\% & 2\% & 5\% & 10\% & 20\% & 50\% & 100\% \\
    \midrule
	Semseg && 23.78 & 28.62 & 36.53 & 39.05 & 43.85 & 47.37 & 50.76\\
	    \midrule
	Semseg + Rot && 22.05 & 26.92 & 36.29 & 39.64 & 43.93 & 48.01 & 50.70\\
	Semseg + MoCo && 25.42 & 30.06 & 37.44 & 40.88 & 44.64 & 48.99 & \textbf{51.41} \\
	Semseg + DenseCL && \textbf{25.96} & \textbf{30.67} & \textbf{38.66} & \textbf{41.40} & \textbf{45.21} & \textbf{49.00} & 50.93\\
    \bottomrule
  \end{tabularx}}
\label{table:sup_coco_semseg_zeroshot}
\end{table*}

\begin{figure*}[t!]
 \centering
 \vspace{-0.2in}
 \subfloat[Depth]{\label{fig_sup:depth_joint_perf}
      \includegraphics[width=0.35\linewidth]{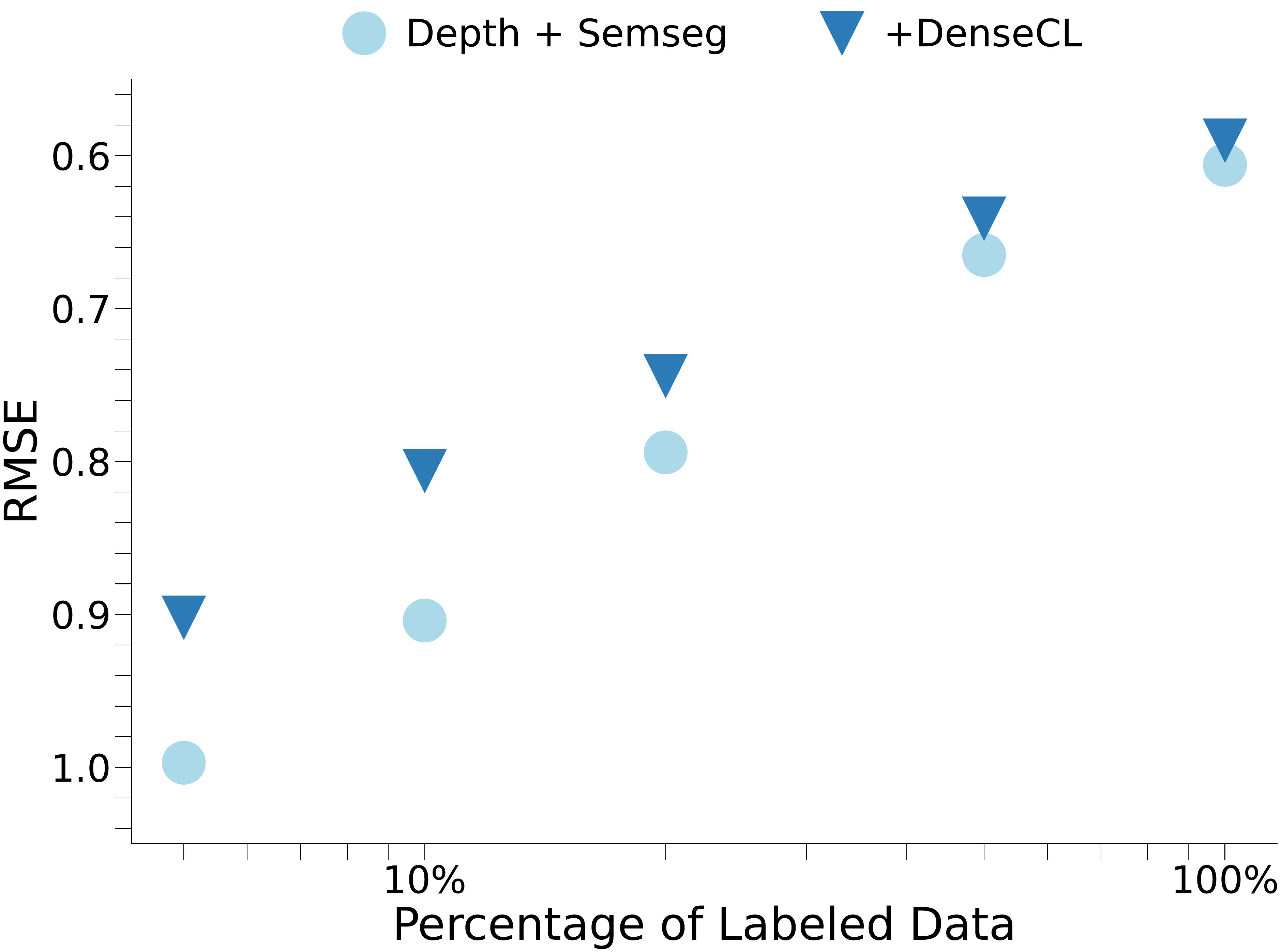}}
      ~
  \subfloat[Semseg]{\label{fig_sup:semseg_joint_perf}
      \includegraphics[width=0.35\linewidth]{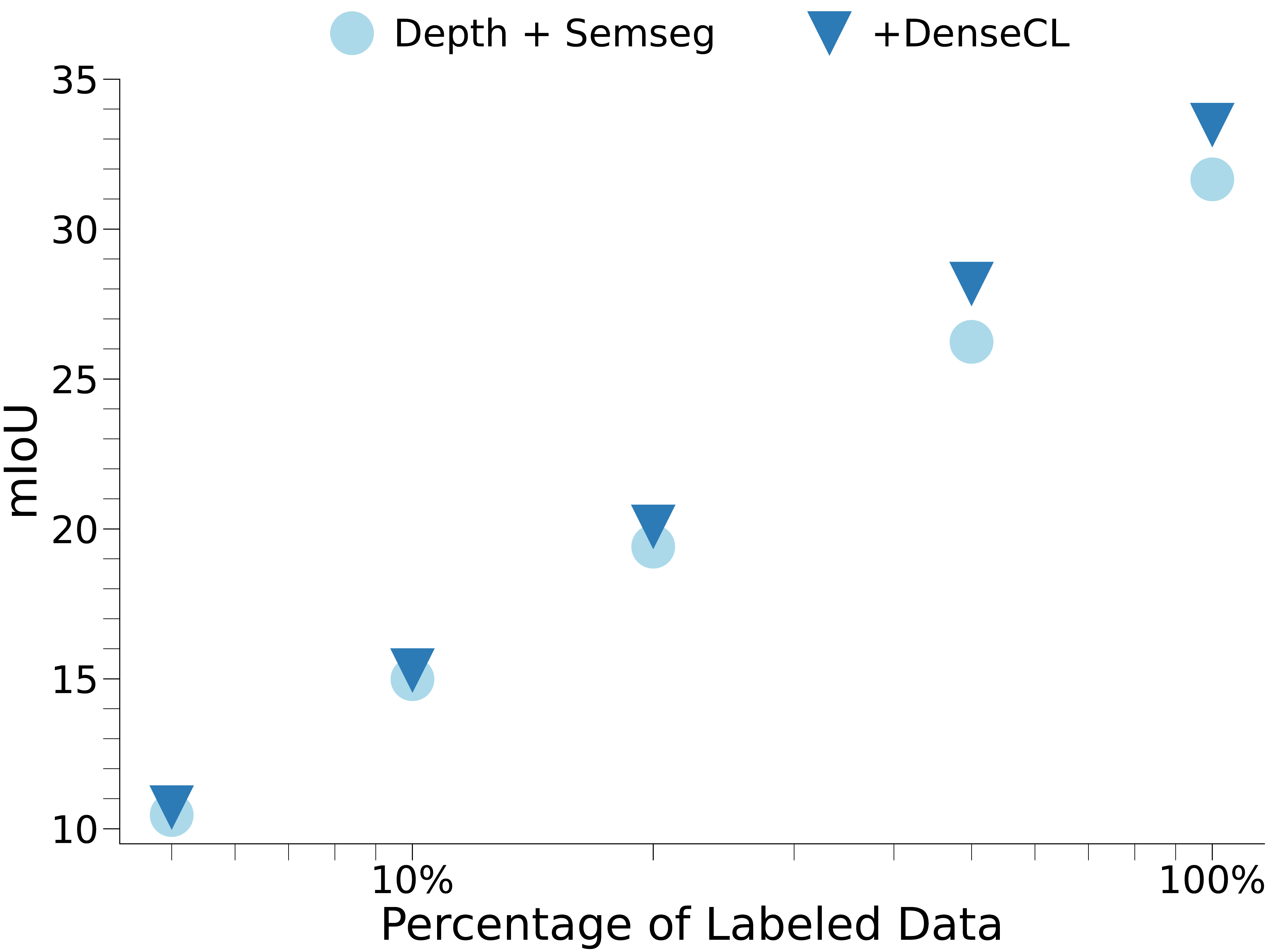}}
\caption{Performance of (a) monocular depth estimation (Depth) and  (b) semantic segmentation (Semseg) on NYUD-v2 for their multi-task model. The multi-task model combined with CompL yields consistent improvements in both tasks.}
\vspace{-0.1in}
\label{fig:sup_mtl}
\end{figure*}

\parsection{Joint optimization}
In 
Table~\ref{table_sup:depth_semseg_joint_perf}
of the main paper, we presented the performance of the baseline multi-task model (Depth + Semseg), and the model trained jointly with DenseCL (Depth + Semseg + DenseCL). 
For ease in comparison between the different models, Fig.~\ref{fig:sup_mtl} additionally visualizes the results.
Training under CompL enhances the performance of both Semseg and Depth, with Depth observing a noticeable gain over Semseg in low data regimes. 
As discussed in the main paper, this can be attributed to the DenseCL hyperparameters being optimized directly for the improvement of Depth.
Furthermore, more advanced loss balancing schemes~\cite{chen2018gradnorm} could yield a redistribution of the performance gains, however, such investigation is beyond the scope of our work.

%% file: text/exp_details.tex
\begin{figure}[t]
 \centering
  \subfloat[Input image]{
      \includegraphics[width=.35\textwidth]{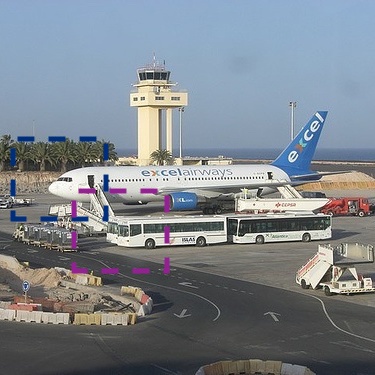}}
\\
  \subfloat[Blue crop]{
      \includegraphics[width=.17\textwidth]{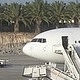}}
~
      \subfloat[Purple crop]{
      \includegraphics[width=.17\textwidth]{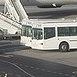}}
\caption{Low overlapping crops can be semantically different. This is more apparent in dense prediction datasets where multiple objects can be present in each image.}\label{fig:segm_crop}
\vspace{3.9in}
\end{figure}

\section{Experiment Details}

\subsection{Codebase}
In this work, we base our experiments on the VIsion library for state-of-the-art Self-Supervised Learning (VISSL)~\cite{goyal2021vissl}, released under the MIT License. VISSL includes implementations of self-supervised methods, and was adapted to enable for the joint optimization of the existing algorithms with supervised methods (semantic segmentation, monocular depth estimation, and boundary detection). The code will be made publicly available upon publication to spark further research in Composite Learning (CompL). 

\subsection{Technical details}

All experiments were conducted in our internal cluster using single V-100 GPUs. Due to the considerable costs associated with multiple runs (beyond our compute infrastructure capabilities), we run all experiments with a random seed of 1, the default setting of VISSL. We provide additional details about different aspects that affect the self-supervised methods below:

\parsection{Memory bank}
MoCo~\cite{he2020momentum} and DenceCL~\cite{wang2020dense} utilized a memory bank to enlarge the number of negative samples observed during training, while keeping a tractable batch size. Specifically, both methods use a memory bank of size 65,536. All the datasets we used in our study are of a smaller magnitude compared to that memory bank, e.g.\ 10,582 and 795 for PASCAL VOC 2012 (aug.)~\cite{hariharan2011semantic} and NYUD-v2~\cite{silberman2012indoor}, respectively. We therefore set the memory bank to have the same size as the training dataset, yielding a single positive per sample, and therefore allowing for the direct use of the InfoNCE loss~\cite{oord2018representation}. A larger memory bank can also be used, however the contrastive loss would need to be adapted to account for multiple positives~\cite{khosla2020supervised}.

\parsection{Image cropping}
We use nearly identical augmentations to those proposed in MoCo v2~\cite{chen2020improved} for the self-supervised methods of \cite{he2020momentum,wang2020dense}, but found it beneficial to modify image cropping. In most classification datasets, each image is comprised of a single object, and thus low overlapping crops can still include the same object. In dense tasks such as semantic segmentation, low overlapping crops can contain different objects (Fig.~\ref{fig:segm_crop}). We follow the practice of~\cite{tian2020makes} and find a constant crop size and distance between the two patches for each task. We empirically find that square crops of size 384 with a distance of 32 pixels on both axis works best for semantic segmentation, crops of size 283\texttimes373 (to maintain input size ratio) with a distance of 8 pixels worked best for depth, and square crops of size 320 with a distance of 4 pixels worked best for boundary estimation. 

\parsection{DenceCL global vs local contrastive}
DenseCL, as discussed in 
Sec.~\ref{ss_methods}
of the main paper, includes a global and a local contrastive term. The importance of the local contrastive term is weighed by a constant parameter. The original paper found that 0.7 for local contrastive and 0.3 for global contrastive performed best for detection, but used 0.5 to strike a balance between the downstream performance on detection and classification. In our study, we also found 0.7 for local contrastive yields the best performance, and as such, used it for all DenseCL experiments.

\parsection{Hyperparameter $\lambda$}
During training, the auxiliary loss is scaled by the hyperparameter $\lambda$, weighting the contribution of the auxiliary self-supervised task.
The hyperparameter $\lambda$ was selected by performing a logarithmic grid search, as commonly done in MTL literature, chosen from the set \{0.05, 0.1, 0.2, 0.5, 1.0\}.
We found the performance of the models to be consistent when $\lambda$ is in the range of 0.1 to 0.5, as seen in Table~\ref{table:sup_semseg_lambda}.
The performance quickly degrades for values an order of magnitude larger as the model prioritizes the auxiliary task over the target task, while smaller values converge to the baseline performance.

\begin{table}[h]
  \centering
   \caption{Ablation of the $\lambda$ parameter for the semantic segmentation model trained jointly with DenseCL. The model yields comparable performance for all three values.}
      \footnotesize
  \begin{tabularx}{0.4\linewidth}{@{}lcXr@{}}
    \toprule
\multirow{2}{*}{$\lambda$} &&    \multicolumn{2}{c}{Labeled Data} \\
    \cmidrule{2-4} 
    && 10\% & 50\% \\
    \midrule
    
	0.1  && 57.21  & 68.64 \\
	0.2 && \textbf{57.33} & \textbf{68.81} \\
	0.5 && 57.27 & 68.79 \\
    \bottomrule
  \end{tabularx}
\label{table:sup_semseg_lambda}
\vspace{4.3in}
\end{table}